# A new approach for extracting the conceptual schema of texts based on the linguistic Thematic Progression theory

Elena del Olmo[1] and Ana María Fernández-Pampillón[2]

**Abstract.** The purpose of this article is to present a new approach for the discovery and labelling of the implicit conceptual schema of texts through the application of the Thematic Progression theory. The underlying conceptual schema is the core component for the generation of summaries that are genuinely consistent with the semantics of the text.

## 1. INTRODUCTION

Automatic Summary Generation was first proposed in the late 1950s. Outstanding examples of this early stage are Luhn (1958), whose method is based on sentence extraction relying on its words weightings, inferred from *TF-IDF* metrics, or Edmundson (1969), who proposed novel sentence weighting metrics, such as the presence of words from a predefined list, the presence of the words of the title of the document or its positioning at the beginning of documents and paragraphs. These are paradigmatic examples of the first extractive summarization techniques: techniques based on the verbatim extraction of the most relevant parts of a text. The generated text summary was, thus, a collection of sentences considered relevant but, often, semantically inconsistent because of the overall weakness in coherence (the text does not make overall sense) and cohesion (the sentences are connected incorrectly). The summary generated was consequently a poorly connected text with no global meaning, presumably due to the assumption of independence of the extracted sentences (Lloret *et al.* 2012).

Currently, five main approaches to extractive techniques can be distinguished: (i), statistical approaches (Luhn 1958, McCargar 2004, Galley 2006), based on different strategies for term counting, (ii), *topic-based* approaches (Edmundson 1969, Harabagiu *et al.* 2005), which assume that several topics are implicit in a text and attempt to formally represent those topics, (iii), graph-based approaches (Erkan *et al.* 2004, Giannakopoulos *et al.* 2008), based on the representation of the linguistic elements in texts judged to be relevant as nodes connected by arcs, (iv), discourse-based approaches (Marcu 2000, Cristea et al. 2005, da Cunha et al. 2007), whose target is to capture the discursive relations within texts, and, (v), machine learning approaches (Aliguliyev 2010, Hannah *et al.* 2014), intended to reduce text summarization to a classification task by assigning a relevance value to each sentence.

Although historically less addressed in the literature, abstractive models try to address the lack of coherence and cohesion in the summaries, using some source of semantic internal representation of the text (which can be merely the output of an extractive process) to generate the summary, composed of sentences not necessarily included in the original text. Although this approaches theoretically improve the consistency issue, they introduce a new complexity layer: a natural language generator module. Despite this, nowadays text summarization research is progressively shifting towards abstractive approaches (Lin *et al.* 2019).

Traditionally, abstractive summarization techniques have been classified into *structure-based,* intended to populate predefined information structures out of the relevant sentences of the texts, and *semantic-based*, involving a wide variety of knowledge representation techniques. Regarding the former, depending on the structural schema chosen, it is possible to identify, (i), tree-based models (Kikuchi *at. al.* 2014), which perform different strategies for syntactic parsing analysis in order to codify paraphrasing information mainly by linking and reducing the syntactic sentence trees of the text, (ii), template-oriented models (Elhadad *et al.* 2015, Wu *et al.* 2018, Wang *et al.* 2019), which rely on extraction rules led by linguistic patterns matching sequences of tokens to be mapped into predefined templates, (iii), ontology-based models (Nguyen 2009, Baralis *et al.* 2013), which are highly domain-dependent and include a hierarchical classifier mapping concepts into the nodes of an ontology, and, (iv), *rule-based* models (Genest *et al.* 2011), based on extraction rules operating on categories and features representative of the content of the text. Regarding the semantic-based techniques, there are interesting approaches based on the concept of *information item* (Gatt *et al.* 2009), the smallest units with internal coherence, in the format of subject-verb-object triplets obtained through semantic role labeling, disambiguation, coreference resolution and the formalization of predication. Besides, there are approaches based on discourse information (Gerani *et al.* 2014, Goyal *et al.* 2016), *predicate-argument* tuples (Li 2015, Zhang *et al.* 2016) and semantic graphs (Liu *et al.* 2019).

The aforementioned tendency towards abstractive approaches in recent years is framed at a stage when Deep Learning models have proved to be particularly promising for using vector spaces as a way to address the shortcomings of discrete symbols as the input for NLP tasks, such as tokens or lemmas, which cannot represent the underlying semantics of the concepts involved. This new paradigm has provided techniques for both extractive and abstractive summarization, such as the clustering of sentence and document embeddings, or the generation of correct sentences given a sentence embedding and a language model. Remarkable examples are the contributions of Templeton *et al.* (2018), who compare different methods of sentence embeddings computing their cosine similarity, or Miller et *al.* (2019), who proposed *k-means* clustering to identify sentences closest to the centroid for summary selection.

[1] General Linguistics department, Complutense University of Madrid, Spain, email: elenadelolmo@ucm.es
[2] General Linguistics department, Complutense University of Madrid, Spain, email: apampi@ucm.es

In addition to the distinction between extractive and abstractive approaches, there is a crucial challenge in automatic summarization which affects them both: the subjectivity of the accuracy scoring of summaries. This implies a new difficulty in the creation of objective gold datasets composed of correct summaries. In this context, unsupervised summary models, such as the one proposed in this paper, which does not require training labelled data, has become particularly relevant. Among the unsupervised approaches we can highlight, (i), approaches which are extensions of word embedding techniques, such as the *n-grams* embeddings (Mikolov *et al.* 2013), or *doc2vec* (Le *et al.* 2014), (ii), the *skip-thought* vectors (Kiros *et al.* 2015), (iii), the *Word Mover's Distance* model (Kusner *et al.* 2015), (iv), *quick-thought* vectors (Logeswaran *et al.* 2018), and, (v), models based on contextual embeddings obtained from transformers, such as SBERT (Reimers *et al.* 2019).

This paper addresses one of the weaknesses of extractive models discussed above: the lack of coherence in the summaries produced, especially when there are insufficient linguistic datasets in a language for applying machine or Deep Learning methods. In this respect, the solution we propose identifies implicit conceptual schemas from texts using the morpho-syntactic knowledge currently provided by NL analyzers. The paper is organized as follows: in section 2 we define the hypothesis and objectives of the research work. In section 3 we present a review of the linguistic theories on which we base our solution. In section 4 we present our solution: the application of both theories for the identification of the text conceptual schema. In section 5 we study the feasibility of our solution for the automatic extraction of thematic progression in Spanish, a language with few linguistic datasets for text summarization. Finally, in section 6 we draw the conclusions of this work and present our future research lines.

## 2. HYPOTHESIS AND OBJECTIVES

Our hypothesis is that applying the Thematic Theory and the Thematic Progression Theory to annotate the discourse features *theme*, *rheme* and their coreferences will allow us to extract *thematic progression schemas*, which represent the implicit conceptual schemas of texts. Therefore, our aim is obtaining an internal representation of the text informational structure as a formal representation for text summarization. The advantage of this solution is that it can be applied to any language regardless of whether or not there are enough training data for the implementation of machine learning and Deep Learning techniques. In our work we will use Spanish as the language to study the feasibility of the solution. We also hope to contribute to the generation of summaries in Spanish, a task currently performed with moderate efficiency due to the limited availability of linguistic resources.

## 3. REVIEW OF LINGUISTIC THEORIES

### 3.1. Thematic theory

The thematic theory is framed within the optics of linguistic analysis corresponding to the informational layer. The uses and applications that the authors have been giving to terms such as *theme*, *focus*, *topic* and notions such as *new information* or *emphasis* have been overwhelmingly numerous (Gutiérrez 2000). In accordance with the Thematic theory, in descriptive and narrative texts, known information, or *theme*, is consensually described to be positioned at the beginning of sentences. By contrast, the phrases containing the informative contribution of the sentence, also known as *rheme*, tend to be located further to the right, ahead in the time of enunciation. This description is consistent with how the acquisition of new knowledge is described at the neurological level, through networking the known with the novel or by altering pre-existing relationships (McCulloch *et al.* 1943). In order to clarify how we will use these concepts, we present here a series of examples adapted from Gutiérrez (2000: 18) and their corresponding answers:

1. *Who joined Luis today? / Luis was joined today by Jo.*
2. *When did Peter join Luis? / Peter joined Luis today.*
3. *Who joined Peter today? / Today Peter joined Luis.*

That these statements are different is a standard judgment for any native speaker. Although they share the same representative function, *i.e.* their syntactic and semantic relations do not differ, they show different informative functions. Therefore, in spite of transmitting the same information about the world, they do not *inform* in the same way. Accordingly, the underlying assumption of our proposal is that the *thematic status* of a phrase (like *who*, *when* and *who* in the examples above, respectively) is relevant in terms of the prevalence of the concept involved for the summarization of a document. Their clustering along a document, taking into account the thematic progressions patterns found, as further explained in the next section, is expected to reveal the conceptual schema of the text.

### 3.2. Thematic Progression theory

Daneš (1974: 114) presents the thematic progression as the choice and arrangement of discourse themes, their concatenation, hierarchy and relation with the topics of texts. Accordingly, he argues that it is possible to infer the informational schema of a text from its theme-*rheme organization*. He proposes three main typologies of thematic progressions: (i), linear progression, in which the rheme of one sentence is the theme of the subsequent sentence, (ii), constant progression, in which a theme is repeated over several sentences, and, (iii), derived progression, in which several topics are derived from the same inferred *hypertheme*. Apart from these three basic types, Daneš (1974: 120) also proposed that the combination of them can lead to thematic progressions of higher levels of abstraction, such as, (iv), the split rheme progression, which consists of the existence of a complex rheme, whose hyponyms and meronyms are themes of the subsequent sentences. Finally, he concludes (1974: 122) that the study of the thematic organization of a text could be useful for numerous practical applications, among which outstands information retrieval, given the performance achieved nowadays by the tools available for the automatic text analysis.

## 4. THEMATIC PROGRESSION AS A MODEL FOR SEMANTIC MODELLING

The usefulness of the thematic or rhematic roles of concepts along texts for automatic text summarization arises from two main facts. On the one hand, the theoretical validation of the concept of thematic progression enjoys consensus among researchers as a relevant description for the semantic structure of texts. On the other hand, although it has been traditionally examined through the optics of the Pragmatics layer, the thematic or rhematic status of a concept is actually embodied in the surface syntactic layer, which is prone to be represented in an easy-to-compute form.

Concerning the correlation between the theme of a sentence and its syntactic structure, crucial for its automatic annotation, Halliday (1985) proposed an interesting categorization based on the concept of linguistic markedness. Thus, declarative sentences of SVO languages, such as English or Spanish, may contain unmarked themes, prototypically subjects preceding main verbs, or marked themes, such as circumstantial attachments, complements, or sentences with predicate constructions. Examples for the former are the first and second sentences of the examples provided above, with *Luis* and *Peter* as unmarked themes respectively, whilst the third sample sentence is an example for the latter, with *today* as theme. Thematic equative sentences, such as *What I want is an advice*, would be excluded from this categorization. For interrogative sentences, the unmarked themes are the definite verbs for *yes-no* questions, such as *did deliver* in *Did the boss deliver a speech*, and interrogative pronouns and similar phrases for non-polar questions, such as *where* in *Where is Berlin located?*, whilst circumstantial adjuncts would be considered marked.

Besides, a constituent which is not the theme of a sentence may appear occasionally in a prototypical thematic position. This phenomenon has been referred to by several names, such as *focussing* (Campos *et al.* 1990), *focus preposition* (Ward 1985) or *thematization* (Chomsky 1972). Examples of this type of informational structure are *It was Pedro who lied to me*. A number of authors (*e.g.* Gutiérrez 2000: 34) have argued that the intent of this information schemas is to gain the attention of the interlocutor to overcome their presumed predisposition to receive information that is at some point contrary to that which is intended to be communicated, or simply to emphasize a certain aspect in the informational process. This nuance of enunciative modality would undoubtedly be applicable for the weighing of the relevant concepts for a proper summary, especially since the syntactic structures involved are relatively easy to formalize.

In short, it is possible to locate the discourse elements *theme* and *rheme* using syntactic knowledge. To the extent that syntactic analysis is a task that can be considered well solved in NLP, it seems feasible to be able to automatically locate the *theme* and *rheme* in every sentence. The next step to obtain the thematic progression schema, *i.e.* the conceptual schema of a text, is to connect each theme and rheme of the text in the ultimate thematic path.

## 5. A FIRST STUDY OF THE FEASIBILITY OF THEMATIC PROGRESSION EXTRACTION IN SPANISH TEXTS

### 5.1. Methodological framework

Our framework for the automatic extraction of the thematic progression schema of a text comprises two sequential phases (figure 1), each one composed of various processing steps.

The first phase requires an input file in the standard CoNLL-U format [3], which basically consists of a text segmented into sentences and their labelled tokens. The linguistic features required for each token, in addition to its identifier and form, are: the lemma, the part-of-speech tag, its morphological features, the type of syntactic-semantic dependency relation in which it is involved and the identifier of its head (see figure 2 as an example). This is a purely symbolic processing phase, which involves the successive application of tree rewriting rules based on the lexical and syntactic information available. Its objective is the automatic annotation of the subtree corresponding to the theme and the subtree corresponding to the rheme for each sentence, by the addition of two new theme and rheme tags in the CoNLL-U input text.

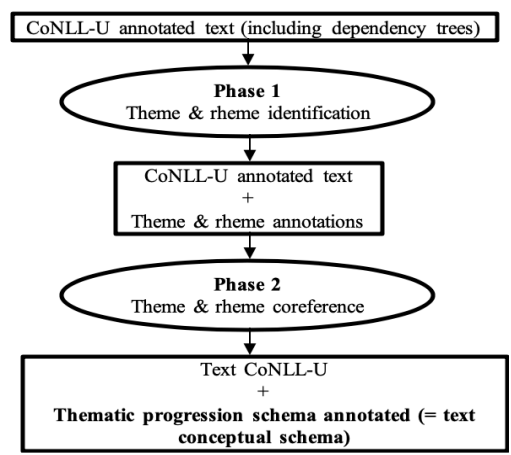

**Figure 1.** Architecture of the thematic progression analyzer

The second phase takes as input the output of the first one, which contains the information concerning the delimitation of the boundary between the theme and the rheme for every sentence. Its objective is the identification and clustering of the themes and the rhemes of the sentences of a text, in such a way that each set gathers the subtrees with a common reference concept. By this means, it would be possible to extract the different thematic progression patterns existing throughout the text and, therefore, to obtain as the final output the progression patterns covered by each concept, *i.e.* the conceptual schema of the text. This sort of output does not constitute by itself a subset of the original text, but it does imply the addition of a layer of supra-segmental information to it. This new information facilitates its readability by enabling the readers to review those parts in which the concept they are interested in first appears (possibly as a rheme of a thematic concept already implicit in the content of the text to that extent), together with other related concepts (*i.e.* the themes of which it is a rheme). Besides, this new layer can also assist the readers in their relevance judgments on the significance of the concepts involved, depending on how many times they are instantiated in thematic positions throughout the text).

One major challenge for the procedural design of this second phase is posed by the fact that the symbolic processing algorithm carried out by the previous phase produced sequences of theme and rheme *subtree phrases* instead of individual theme and rheme *tokens*. Phrase (or sentence) clustering requires some sort of semantic similarity computing. Unfortunately, the field of semantic similarity for sentences and phrases has not been the focus of extensive investigation until quite recently. Nonetheless, promising new insights have emerged during the last few years, strongly motivated by advances in the area of Vector Space Semantics and the prevalence of *tranfer learning*. In this regard, we are considering the application of SBERT (Reimers *et al.* 2019) to work out phrase similarity of textual themes and rhemes. SBERT is a modification of the BERT network to derive semantically meaningful sentence embeddings. BERT (Devlin *et al.* 2018) was designed to pre-train deep bidirectional representations from raw text, which can be easily fine-tuned (by adding task-related output layers) in order to create state-of-the-art models for different tasks.

---
[3] https://universaldependencies.org/format.html

## 5.2. Preliminary study on the feasibility of the automatic identification and annotation of themes in phase 1

Aiming to conduct a first study to verify the feasibility of the phase 1 automatic identification and annotation of textual themes task, we carried out an exploratory corpus survey with Spanish descriptive texts. Our hypothesis is that by using syntactic markers we can infer the pragmatic thematic layer of sentences. The aim for this particular study is testing to what extent it would be effective to annotate preceding subjects as themes in Spanish, since it is a *SVO* language (with a predominant subject-verb-object constituent order), with preceding subjects with a default thematic role, that is, depicting known information within the sentence informational flow.

To test the hypothesis, we have carried out a quantitative and qualitative study with the following steps: (i) first, we analyzed the mean ratio and the ratio per text of preceding subjects, since they are the prototypical themes in Spanish, and, (ii) second, on the basis of this quantitative results, we categorized the casuistry of sentences without preceding subjects, as well as complex preceding subjects, as those containing embedded subordinate clauses or involving coordinate structures. The categorization of complex preceding subjects will allow us to explore ways to simplify them minimizing the information loss, a mandatory step to effectively achieve the identification of theme and rheme coreference in phase 2.

The study has been conducted using the AnCora Surface Syntax Dependencies corpus[4] *(AnCora GLiCom-UPF 1.1)*, published in 2014 at Pompeu Fabra University. It contains 17,376 sentences manually annotated with the lemmas, the *PoS* tags plus other morphological features and both dependency heads and relations for every token. In order to perform the analysis, we designed a symbolic rule-based grammar using subtree extraction operations from the *dependency tree* of the sentences. The theme analyzer scrips developed to process the grammar and to generate the output are publicly accessible from github[5].

As mentioned above, for the grammar definition, an approach based on dependency tree transformations has been applied. Thus, for the task of finding subjects in SVO languages, such as Spanish, two categories of rules were defined:

1. Matching rules to select a child node or a child dependency subtree from a given feature-constrained head node. These type of rules consists of a predicate of arity two, with the name of the dependency relation as a functor (or predicate name) with a first argument as the selected feature-constrained parent node and a second argument with one out of two subtree matching options: *ALL* if all the children subtree should be selected and *ONE* if only the immediate child node should be selected. For example, as it is shown in figure 2 (obtained from the Freeling 4.1 demo[6]), the *SBJ(deprel:ROOT, ALL)* rule matches the *SBJ* (*i.e.* non-clausal subject) child subtree of a node annotated with a *ROOT* dependency.

2. Matching rules to select a head node or a head dependency subtree from a given feature-constrained child node. This type of rules consists also of the name of a dependency relation as the name of a predicate of arity two, whose arguments are the same as for the child tree selection rules but in the opposite order (embracing the convention by which the first argument is the head and the second one is its child). A prominent context of application for this type of rule is sentence simplification when several propositions are involved in either the theme or the rheme (*e.g.* to transform the sentence *When he had the chance, the chairman declined to appear* into the subtree *The chairman declined to appear*).

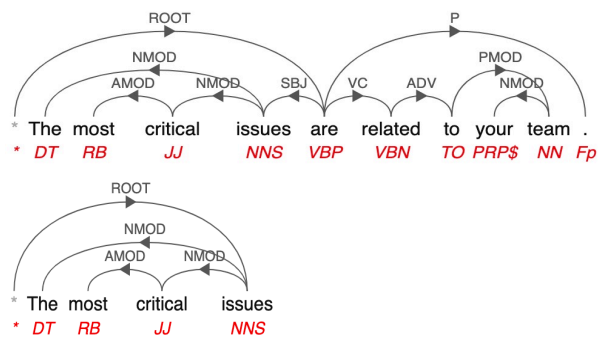

**Figure 2.** Input sentence and output (subtree) of a rule of the first type (Source: Freeling 4.1 demo[7]).

Regarding the input format, the theme analyzer accepts various types of text files and transforms them into the universal *CoNLL-U* text format, where the additional theme and rheme features are added for every token or phrase in the main proposition of sentences. We found that, with our first version of the rules, the theme and rheme automatic annotations were correct in roughly half of the cases, as shown in table 1. Through a careful manual review of the data, these results have enabled us to confirm a significant correlation between the syntactic-semantic and discourse layers outlined by the Thematic theory, and, consequently, the feasibility of the automating identification of, at least, half of the themes.

---

[4] http://clic.ub.edu/corpus/es/ancora-descarregues
[5] https://github.com/eelenadelolmo/HI4NLP/tree/master
[6] http://nlp.lsi.upc.edu/freeling/demo/demo.php
[7] http://nlp.lsi.upc.edu/freeling/demo/demo.php

**Table 1.** Ratio of preceding subjects in AnCora corpus.

|  | GLiCom-UPF 1.1 | Freeling version |
|---|---|---|
| Sentences with preceding subjects as themes | 50.4 % | 46.2 % |

The qualitative analysis of exceptions, which are the unmatched sentences, *i.e.* sentences without preceding subjects or with very complex preceding subjects, revealed a strong presence of thematization as the most relevant finding. Thematization has been referred to by the Thematic theory as a relevant discourse feature, indicating a break in the information flow of the text. A promising conclusion of our analysis of the thematization patterns found is their suitability for being matched with the rule formalism designed.

Another interesting finding obtained from the qualitative analysis of the manually annotated version revealed a strong presence of subordinate clauses, mostly within rhemes. This implies the necessity of incorporating more complex rules in the grammar in order to select the most informative and simplest proposition for every sentence. The observed patterns have been categorized into three main categories (the most informative clauses appear in bold and the themes selected by this first version of rules are underlined):

1. Sentences whose root clause is the most relevant (*e.g. Since pharmacists work with a high profit margin, **the business opportunity** is huge*).
2. Sentences whose root clause is not the most relevant. (*e.g. The main factor is that **electricity consumption during the summer is now not much lower than it used to be***).
3. Sentences whose root clause is not the most relevant but provides a crucial modality feature for information retrieval (*e.g. The bosses are convinced that **John deliberately cut it***).

Another result drawn from the qualitative analysis is the insufficiency of the rule language designed for our preliminary study in its current state, since it has been proven not to be sufficiently expressive to cover all the necessary operations observed. The most important missing feature is the capacity to declare node or arc variables which could be instantiated in subsequent rules. This would allow to select more sophisticated subtree patterns, *e.g.* those involving multiple dependencies on branches indirectly linked to the matched node. Furthermore, although the implementation would not entail any major complications in this case, it would also be worthwhile to develop some convention for the definition of lexicons allowing the rules to be parameterized, primarily for the sake of simplicity and clarity in the codification of the rules.

Additionally, in order to ensure the generality of the grammar, the matched themes obtained with the original corpus were compared with the corresponding results in a new version of the corpus automatically annotated with the *Freeling analyzer*, which constitutes the actual usage scenario for Spanish unannotated texts. As the figures in table 2 suggest, a pervasive tendency for undermatching thematic preceding subjects has been confirmed in the Freeling version. Through careful examination, we found that the vast majority of actually mismatched annotations involve some type of coordinated or juxtaposed clauses. These syntactic structures are analyzed by Freeling with a highly fluctuating dependency structure, which is quite different from the analysis in *GLiCom-UPF 1.1*. This high variability in the syntactic tree accounts for the vast majority of both the undermatched and overmatched preceding subjects. Therefore, it becomes clear that it is necessary to adapt the rules according to the analyzer chosen. Such a procedure is not a trivial task, since the patterns of automatic annotation of certain types of structures are highly variable but, nevertheless, it is manageable by means of including fine-grained rules.

**Table 2.** Ratio of suspected annotation errors.

|  | Freeling version |
|---|---|
| Suspected overmatches | 1283 (7.2 %) |
| Suspected undermatches | 3550 (20.1 %) |

## 6. CONCLUSIONS AND FUTURE WORK

As observed in the study conducted, this first approach to rule-based theme annotation seems to claim the theoretically hypothesized correlation between the syntactic-semantic and discourse layers required by our proposal. However, the qualitative analysis of sentences with and without preceding subjects revealed the need for more complex tree rewriting rules to achieve a more accurate theme selection in order to obtain thematic progression schemas from texts.

To implement a formal tree rewriting language with more expressive power than the prototype presented, we are currently exploring the embedding of Grew (Bonfante *et al.* 2018), a purpose-specific graph rewriting rule language, into the first module of the framework proposed. Grew rules address the expressive power issues discussed above. Although Grew rules suffer from certain weaknesses, such as the absence of recursive subtree annotation options and the loss of ordering in nodes, we are complementing it with additional functions to provide it with the necessary expressive power to handle the most complex scenarios for theme and rheme selection.

Regarding subordination, *i.e.* sentences with several propositions with different syntactic status, we are currently working on two feasible options for sentence simplification: (i), the choice of the most relevant proposition for every sentence, and, (ii), the choice of the ordered subset of its *n* more relevant clauses. In addition, this study shows the necessity to implement an algorithm to infer the modality from the main verb, since reported speech and subordinate clauses implying opinions are very frequent in journalistic texts. In order to ensure the objectivity in the conceptual schemas, the modal status of its components must be taken into account within their coreference chains. Finally, especially in sentences containing subordinate clauses, whether or not they involve modality issues, our study revealed that it would be useful to design a separate convention to simplify the implementation of lexicon-based rules in order to capture lexical semantic generalizations.

## ACKNOWLEDGEMENTS

We thank the anonymous reviewers for their useful suggestions. This work has been supported by the Research Project TIN2017-88092-R.